\documentclass[10pt,twocolumn,letterpaper]{article}

\usepackage[pagenumbers]{wacv} 

\usepackage{graphicx}
\usepackage{amsmath}
\usepackage{amssymb}
\usepackage{booktabs}
\usepackage{xspace} 
\usepackage{booktabs}
\usepackage{color, colortbl}
\usepackage{multirow}
\usepackage[accsupp]{axessibility}
\usepackage[pagebackref,breaklinks,colorlinks]{hyperref}
\usepackage[dvipsnames,xcdraw]{xcolor}
\usepackage{ifthen}
\usepackage[capitalize]{cleveref}
\crefname{section}{Sec.}{Secs.}
\Crefname{section}{Section}{Sections}
\Crefname{table}{Table}{Tables}
\crefname{table}{Tab.}{Tabs.}

\def\wacvPaperID{2109}
\def\confName{WACV}
\def\confYear{2024}

\newboolean{qc}
\setboolean{qc}{false}
\begin{document}
\newcommand{\todo}[1]{\textcolor{red}{[#1]}}

\iftrue     
\newcommand{\dimpp}[1]{\textcolor{blue}{[DP: #1]}}
\newcommand{\vicky}[1]{\textcolor{magenta}{ \textbf{#1}}}
\newcommand{\vickyc}[1]{\textcolor{magenta}{[VK: #1]}}
\newcommand{\vic}[1]{\textcolor{magenta}{ \textbf{#1}}}
\newcommand{\vicc}[1]{\textcolor{magenta}{[VK: #1]}}
\newcommand{\thanos}[1]{\textcolor{orange}{[AT: #1]}}
\newcommand{\mitsos}[1]{\textcolor{blue}{ \textbf{#1}}}
\else
\newcommand{\dimpp}[1]{\textcolor{blue}{\noindent}}
\newcommand{\vicky}[1]{\textcolor{magenta}{\noindent}}
\newcommand{\thanos}[1]{\textcolor{orange}{\noindent}}
\fi

\newcommand{\mypar}[1]{\vspace{0mm}\noindent\textbf{#1}}

\newcommand{\myparb}[1]{\vspace{0mm}\noindent\textbf{#1}}

\newcommand{\norm}[1]{\left\lVert#1\right\rVert}

\newcommand{\video}{$V$\xspace}

\newcommand{\selframe}{$f_*$\xspace}
\newcommand{\anntype}{$a_{f_*}$\xspace}
\newcommand{\dataset}{MOSE-long\xspace}

\newcommand{\qnet}{QNet\xspace}
\newcommand{\qualdb}{FMQD\xspace}
\newcommand{\mname}{EVA-VOS\xspace}
\newcommand{\jandf}{$\mathcal{J}\&\mathcal{F}$\xspace}


\newcommand{\maskc}{Gray!30}
\definecolor{clickc}{HTML}{9ecae1}

\title{Learning the What and How of Annotation in Video Object Segmentation}

\author{Thanos Delatolas$^{1,2}$ \quad 
Vicky Kalogeiton$^3$ \quad 
Dim~P.~Papadopoulos$^{1,2}$ \quad \\
$^{1}$\,Technical University of Denmark \quad
$^{2}$\,Pioneer Center for AI\\
$^{3}$\,LIX, Ecole Polytechnique, CNRS,  Institut Polytechnique de Paris 
\\
{\tt\small atde@dtu.dk, vicky.kalogeiton@lix.polytechnique.fr, dimp@dtu.dk}
\\\url{https://eva-vos.compute.dtu.dk/}
}

\maketitle

\begin{abstract}

   Video Object Segmentation (VOS) is crucial for several applications, from video editing to video data generation. Training a VOS model requires an abundance of manually labeled training videos. The de-facto traditional way of annotating objects requires humans to draw detailed segmentation masks on the target objects at each video frame. This annotation process, however, is tedious and time-consuming. To reduce this annotation cost, in this paper, we propose \mname, a human-in-the-loop annotation framework for video object segmentation. Unlike the traditional approach, we introduce an agent that predicts iteratively both which frame (``What'')  to annotate and which annotation type (``How'') to use. Then, the annotator annotates only the selected frame that is used to update a VOS module, leading to significant gains in annotation time.
   We conduct experiments on the MOSE and the DAVIS datasets and we show that:
   (a) \mname leads to masks with accuracy close to the human agreement 3.5$\times$ faster than the standard way of annotating videos;
   (b) our frame selection achieves state-of-the-art performance;
   (c) \mname yields significant performance gains in terms of annotation time compared to all other methods and baselines.
\end{abstract}

\section{Introduction}
\label{sec:introduction}
Video object segmentation (VOS) is the task of segmenting and tracking objects of interest in videos~\cite{on_1, on_3,prop_3,prop_7,matching_2, matching_3, matching_6_swem,matching_4_stm,stcn,mivos,aot,deaot,xmem,xmem_plus}.
VOS is a central task for video understanding and enables various applications including video editing~\cite{v_editing_1, v_editing_2}, video synthesis~\cite{vs_1,vs_2}, and video decomposition~\cite{v_decomposition_1}.
Training a VOS model requires videos in which the target objects have been manually annotated with object segmentation masks~\cite{on_1, matching_4_stm, matching_6_swem, stcn, mivos, xmem, xmem_plus,aot,deaot}.
%
This process is expensive and labor-intensive as it requires humans to manually draw a mask at each video frame, which requires 80 seconds per object per frame~\cite{coco}. For instance, annotating only one object in a short 10-second video would require more than 5 hours.
The resource-intensive manual annotation bottlenecks the feasibility of building large-scale VOS datasets, indispensable for training effective models. In turn, this restricts the democratization of annotated video data, thus limiting advances in video understanding.

\begin{figure}[t]
  \centering
  \includegraphics[scale=0.25]{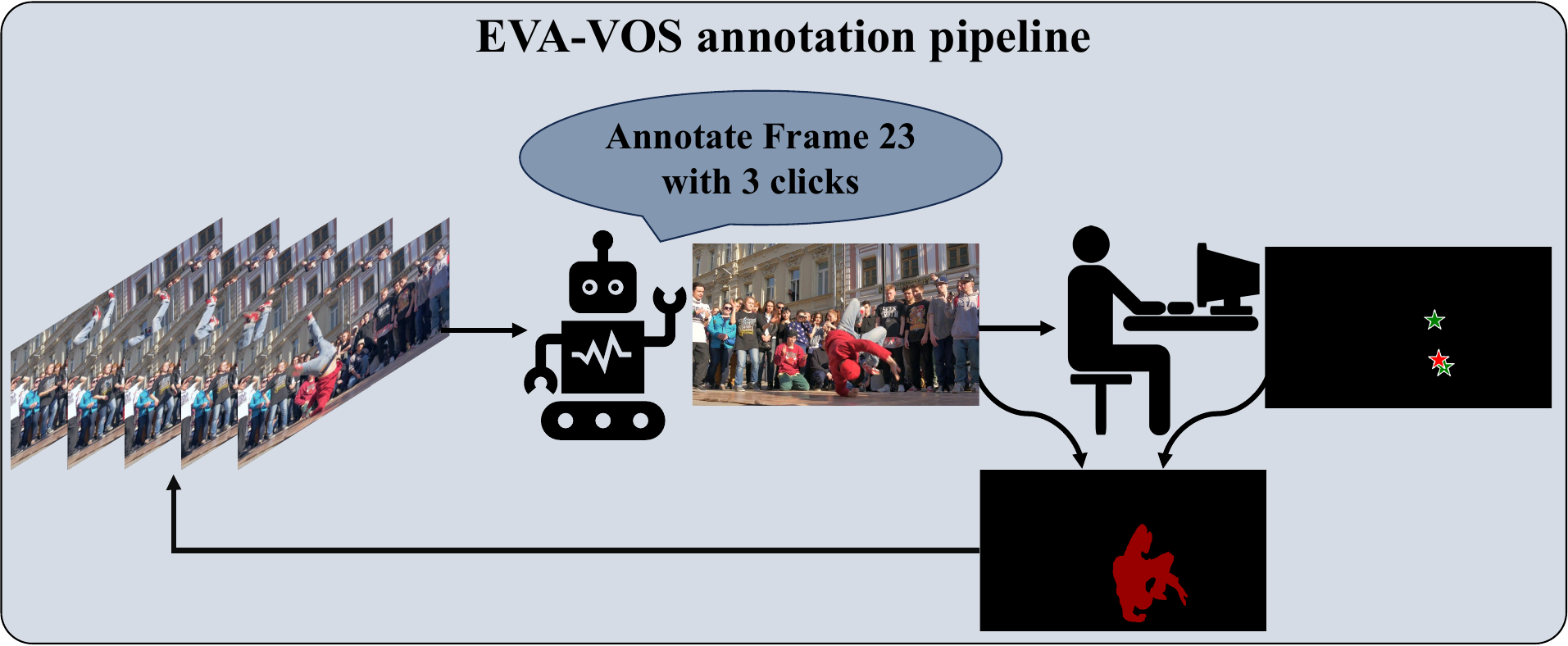}
   \caption{\small \textbf{\mname.} 
   In contrast to the traditional way of annotating objects in videos, 
   we propose to use a human-in-the-loop approach. We introduce an agent that selects the frame that should be annotated (``What to annotate?'') and the annotation type (e.g. clicks, object-mask) (``How to annotate?''). Then, we use the weak annotation to predict a mask for the frame and we propagate it to predict masks for the whole video.}
   \label{fig:teaser}
   \vspace{-2mm}
\end{figure}

To address these limitations, the research community has turned to two solutions.  First, sparsely annotating large VOS datasets~\cite{youtube_vos,ovis,mose,visor,uvo}, and secondly, accelerating the annotation process. Regarding the former, the
 standard way of annotating a VOS dataset~\cite{davis_16, davis_17, youtube_vos, uvo, visor, vost, ovis, mose} 
starts by selecting a subset of frames via a uniform sampling (usually 1-5 fps). Then, these frames are manually annotated by humans who draw a mask at each selected frame. In some datasets, such as VISOR and UVO~\cite{visor,uvo}, the sparsely annotated masks are interpolated to predict dense annotation for all frames. 
However, relying solely on such sparsely annotated large-scale datasets may introduce limitations, especially for applications that demand fine-grained and accurate segmentations throughout the entire video. 

For the latter, i.e., to minimize the annotation cost, the common strategy is interactive segmentation using faster annotation types, such as clicks, scribbles, or bounding boxes. Even though many approaches were proposed for still images~\cite{acuna18cvpr,benenson19cvpr,castrejon17cvpr,li18cvpr,ling19cvpr,papadopoulos17iccv,papadopoulos21iccv} that led to the creation of a larger image segmentation dataset~\cite{benenson19cvpr},
there is limited work in the video domain~\cite{davis_18, chen2020scribblebox,chen2018blazingly,dai2021video}. 
The most relevant to our work is Caelles et al.~\cite{davis_18} who propose a human-in-the-loop interactive VOS. 
The annotator provides a scribble at a frame and a VOS method predicts a mask for each frame. Then, the annotator iteratively selects the frame with the worst segmentation quality and provides scribbles. However, despite its innovation, this approach has two main limitations. 
First, it is unrealistic as the annotator can not identify the worst frame, and even if they could, it would require significant time~\cite{GIS-RAmap}, which defeats the purpose of minimizing the cost. 
%
Second, in challenging frames, low-cost annotation types (e.g. scribbles or clicks) are insufficient to create a good mask, and drawing the full mask is required. 

To overcome these limitations, we propose 
\textbf{\mname}, a human-in-the-loop pipeline (Fig.~\ref{fig:teaser}). Our contribution is the introduction of an agent that predicts iteratively which \emph{frame} should be annotated (frame selection:``What to annotate?'') and which \emph{annotation type} should be used (annotation selection:``How to annotate?'').
Our agent is trained to maximize the annotation impact on the segmentation quality while minimizing the annotation cost.

For the frame selection, we train a model to regress the quality of a segmentation mask. Then, we select the frame that has the maximum distance from its closest pre-annotated frame. 
%
For the annotation selection, we train a deep RL policy that selects an annotation type (action) by maximizing the fraction of the segmentation quality improvement over the annotation time of the annotation type (reward).
Our pipeline iterates between (a) selecting the next frame for annotation and the optimal annotation type, (b) asking annotators to improve a segmentation mask, and (c) predicting new object masks for all frames (Fig.~\ref{fig:pipeline}).




To evaluate our method, we conduct experiments on the MOSE~\cite{mose} and DAVIS~\cite{davis_17} datasets. We first evaluate each stage of the agent (frame and annotation selection) independently and then we show the final results of our full pipeline. Our results show that 
(a) \mname leads to masks with accuracy close to the human agreement 3.5$\times$ faster than the standard way of annotating a VOS dataset;
(b) Our frame selection method achieves state-of-the-art performance;
(c) \mname yields significant performance gains in terms of annotation time compared to other strong baselines.

\section{Related Work}
\label{sec:relwork}
\paragraph{Semi-Supervised Video Object Segmentation (VOS).}VOS methods aim to segment a specific object throughout a video given the object mask in the first frame. Existing VOS methods can be divided into three categories: online fine-tuning~\cite{on_1,on_2,on_3, on_4, on_5, on_6}, propagation-based~\cite{prop_1,prop_2,prop_3,prop_4,prop_5,prop_6,prop_7,aot,deaot} and matching based~\cite{matching_1,matching_2,matching_3,matching_4_stm,matching_5,matching_6_swem,stcn,xmem} methods. Online fine-tuning methods overfit on the object mask of the first video frame. Propagation-based methods use the mask of the previous frame to generate the mask of the current frame. This involves progressively passing on features of the target object from one frame to the next. Matching-based algorithms store features of the target object and classify each pixel of the current frame using similarities in the feature space. Current state-of-the art methods: STCN~\cite{stcn}, XMem~\cite{xmem}, AOT~\cite{aot}, DeAOT~\cite{deaot} achieve remarkable results in the traditional VOS benchmarks (e.g. DAVIS~\cite{davis_16,davis_17} and YouTube-VOS~\cite{youtube_vos}). However, their performance drops notably on new challenging benchmarks: VISOR~\cite{visor}, MOSE~\cite{mose} and VOST~\cite{vost}. 
This is because these datasets contain severe occlusions, disappearance/reappearance of objects, and object transformations. Instead, our method identifies where these methods fail and provides extra annotations to refine the results.

\myparb{Interactive VOS (iVOS)} tackles the human-in-the-loop setting, where the annotator provides quick input (e.g. scribbles~\cite{davis_18,chen2020scribblebox}, points~\cite{chen2018blazingly,clicks_2}, text~\cite{davis_rvos, urvos,openvis}) to a VOS method instead of detailed object masks as in semi-supervised VOS. 
At each annotation round, the annotator selects the frame with the worst segmentation quality and provides a scribble to refine the output mask. The scribble serves as the input to the VOS method, and the process continues iteratively. ScribbleBox~\cite{chen2020scribblebox} builds on top of iVOS by including a prior step where a tracker~\cite{wang2019fast} predicts a bounding box for each frame and then the annotator inspects and refines them. 
We argue that iVOS is unrealistic as the annotator is unable to find the worst frame and even if they did, they would need a significant amount of time, which defeats the purpose of iVOS.

\myparb{VOS dataset annotation.} Manually annotating object masks is time-consuming.
Annotating a VOS dataset requires per frame object masks, linking object masks over time, and quality assurance~\cite{davis_16, davis_17, youtube_vos, uvo, visor, vost, ovis, mose}. As a result, VOS datasets have fewer annotations than image datasets. For instance, the densely annotated DAVIS~17~\cite{davis_17} VOS dataset contains only 13K annotations, vs the 500K annotations in COCO~\cite{coco}. Youtube-VOS~\cite{youtube_vos} scaled up the number of annotations to 197K by annotating every 5 frames. UVO~\cite{uvo} consists of 200K annotations at 1 fps for training and 30 fps for validation. The sparsely annotated masks are interpolated using STM~\cite{matching_4_stm} to cover all frames, and annotators correct the interpolated masks. Similarly, VISOR~\cite{visor} has 271k annotations. Finally, OVIS~\cite{ovis} and MOSE~\cite{mose} are annotated every 5 frames without any interpolation resulting in 296K and 431K masks, respectively. 
These datasets have a similar time-consuming annotation pipeline, constituting a bottleneck. Instead, our method improves this by selecting both the frame and the annotation type (box, clicks, mask) and significantly reduces the cost.

\myparb{Segment Anything (SAM)} was recently introduced~\cite{SAM} and it immediately inspired many new methods~\cite{sam_meets_videos,sam_track,med_sam,cap_anything,av_sam,polyp_sam,sam_pt}. SAM was trained on SA-1B~\cite{SAM} which is the largest dataset for image segmentation as it consists of over 1 billion masks. 
Track Anything (TAM)~\cite{sam_meets_videos} uses SAM and XMem~\cite{xmem} to do interactive VOS with clicks. In particular, the initial mask at the target object is generated from SAM using click prompts and XMem tracks the target object throughout the video. If the quality of the output masks from XMem is low, either SAM is prompted with clicks that are extracted from the affinities of XMem or the annotator prompts SAM to refine the segmentation mask that will be propagated by XMem. Our method differs from TAM because it finds both the frame and the annotation type that SAM will be prompted with to maximize the performance of the propagation while minimizing the annotation cost.

\myparb{Frame Selection in VOS.} The goal of this task is to find a set of frames for the annotator to annotate that maximizes the overall video segmentation quality. BubbleNets~\cite{bubble_nets} predict only the initial frame instead of always using the first frame, which is the standard in the field. GIS-RAmap~\cite{GIS-RAmap}, XMem++~\cite{xmem_plus} and IVOS-W~\cite{IVOS-W} iteratively predict a frame at each annotation round of the iVOS setting. Both GIS-RAmap~\cite{GIS-RAmap} and XMem++~\cite{xmem_plus} are VOS models that first segment frames and then predict the next frame for annotation. GIS-RAmap~\cite{GIS-RAmap} uses the pixel-wise scores of each frame, while XMem++~\cite{xmem_plus} uses the key features of each frame and all previously annotated frames to predict the next frame. IVOS-W~\cite{IVOS-W} uses reinforcement learning (RL) and is also the most closely related work to ours, as it is not based on any VOS model. Both IVOS-W and our method regress the quality of each frame to predict the next one. However, IVOS-W assumes explicit information about the target object because it extracts the region of the image around it~\cite{xmem_plus}. Instead, our method uses the entire frame and regresses its quality. In Sec.~\ref{sub:frame_selection_experiments}, we modify IVOS-W~\cite{IVOS-W} to work for different annotation types other than scribbles, and compare it to our method. To the best of our knowledge, our method (\mname) is the first in the VOS field that predicts the annotation type for each frame.

\section{Method}
\label{sec:method}
\begin{figure*}
  \centering
    \begin{center}
    \includegraphics[width=\linewidth]{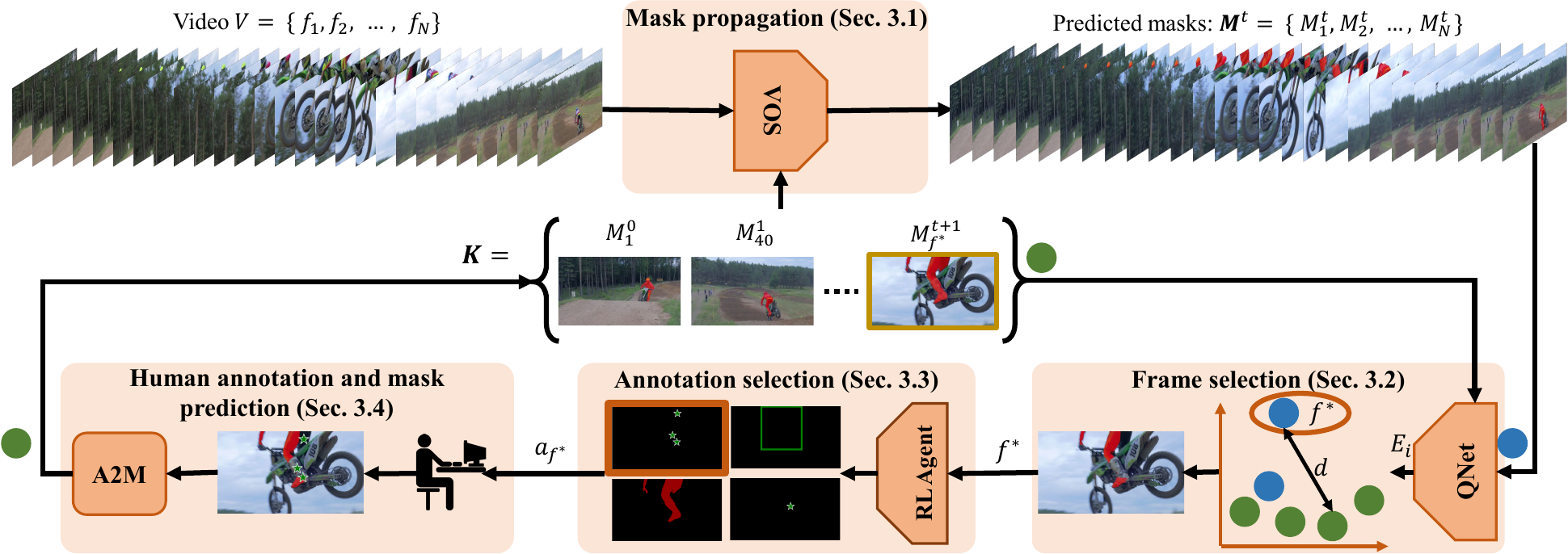}
    \end{center}
    \caption{\small \textbf{\mname}. At each iteration $t$, Mask propagation (Sec.~\ref{sub:mask_prop}) receives a video \video of $N$ frames and a set $K$ containing all previously annotated frames to predict a new set of masks $\mathbf{M}^t=\{M_1^t, M_2^t, \hdots, M_N^t\}$ for all frames. Subsequently, the Frame selection (Sec.~\ref{sub:frame_selection}) stage selects the frame \selframe that should be annotated given the video \video, the predicted masks $\mathbf{M}^t$ and all previously annotated frames $K$. The Annotation selection (Sec.~\ref{sub:annotation_selection}) takes as input the selected frame \selframe and its corresponding mask to predict the most suitable annotation type \anntype. Finally, in the Human annotation and mask prediction (Sec.~\ref{sub:mask_pred}) stage, the annotator interacts with \selframe using the annotation type \anntype and A2M (in this work, we use SAM~\cite{SAM}) predicts the new mask $M_{\text{\selframe}}^{t+1}$ of the frame \selframe, which is then added to the set $K$. 
    }
    \label{fig:pipeline}
\end{figure*}

In this work, we propose \mname, a human-in-the-loop pipeline to annotate videos with segmentation masks using as little annotation as possible (Fig.~\ref{fig:pipeline}). \mname consists of four stages: (a) Mask Propagation (Sec.~\ref{sub:mask_prop}), (b) Frame Selection (Sec.~\ref{sub:frame_selection}), (c) Annotation Selection (Sec.~\ref{sub:annotation_selection}), (d) Annotation and Mask Prediction (Sec.~\ref{sub:mask_pred}).

More formally, at each iteration $t$, the mask propagation receives the input video $V = \{f_1,f_2, \ldots, f_N\}$ of $N$ frames and a set $K$ containing all previously annotated frames to predict a new set of masks $\mathbf{M}^t=\{M_1^t, M_2^t, \hdots, M_N^t\}$ for all frames. Then, the frame selection determines the frame \selframe that should be annotated given $V$ and $\mathbf{M}^t$. The annotation selection determines the most suitable annotation type \anntype from a pool of candidate annotation types $A = \{a_1, a_2, \ldots, a_L\}$. For annotation types, we consider both the case where the annotator manually draws a complete mask (\emph{`mask drawing'}), and the case of weak annotations, where the human intervention is much faster, e.g., \emph{`corrective clicks'}, \emph{`bounding boxes'}, \emph{`scribbles'}, etc.
Finally, the annotator annotates \selframe with \anntype, and the annotation is passed on to the mask prediction, where a new mask $M_{\text{\selframe}}^{t+1}$ is predicted and added to $K$. Note that at $t=0$, the annotator selects the target object and draws a mask on $f_1$.

\subsection{Mask propagation}
\label{sub:mask_prop}
At this step, we predict a set of masks $\mathbf{M}^t$ for all frames using all annotated masks from the set $K$. For this, we use a pre-trained VOS~\cite{mivos} module which takes as input the video $V$ and the masks $K$ and predicts a mask $M_i$ for each frame.

\subsection{Frame selection} 
\label{sub:frame_selection}
Given a video $V$, the predicted masks of each frame $\mathbf{M}^t$, and the set $K$ containing all previously annotated frames, our aim is to find the frame to be annotated \selframe at iteration $t$ in order to have the highest improvement of the video segmentation quality at iteration $t+1$. Annotating \selframe will enhance the segmentation quality of the video $V$ via mask propagation, and as a result, we want to select the frame that will have the most impact on the mask propagation stage (Sec.~\ref{sub:mask_prop}). 
Intuitively, we want to select frames that maximize the diversity among the selected ones and at the same time have low segmentation quality so that we maximize the impact on the final performance.
We first train a model to assess the segmentation quality of each frame, and then we use the learned frame representations to select \selframe.

\myparb{Architecture.}
To assess the mask quality of each frame, we introduce the Quality Network (\qnet) which takes in a frame $f_i$ and its corresponding mask $M_i^t$ and performs mask quality classification into $B$ classes, where $0$ represents the worst quality and $B-1$ the best. The value of $B$ 
determines the number of bins in which the segmentation quality is divided. \qnet consists of two image encoders~\cite{resnet} in parallel branches, one for the frame $f_i$ and one for the mask $M_i^t$. The embeddings from each encoder are then concatenated and fed into a linear classifier with $B$ outputs.

\myparb{Training.} We train \qnet in a supervised way with a cross-entropy loss on a simulated training set. To generate a realistic training set, we simulate a number of iterations with \mname (Fig.~\ref{fig:pipeline}); at each iteration, we compute the segmentation quality of each frame and assign a quality label to each mask $M_i^t$. We simulate our training set with random and oracle selections at each iteration as follows: A random selection chooses \selframe randomly, excluding frames in $K$, while an oracle selection chooses the frame \selframe with the worst segmentation quality.

\myparb{Selected frame \selframe.} 
We select \selframe as the one with the maximum distance in the feature space from its closest previously annotated frame. To this end, we first extract embeddings $E_i$ from \qnet.
Next, we compute the L2 distance between each embedding of a frame $j$ in $K$ and all frames of $V$. 
Finally, we assign the minimum distance to each embedding $i$, and we select the frame with the maximum distance. This process can be mathematically described as: 
\begin{equation}
  f_* = \underset{i \in \{1,2\ldots N\}}{\arg\max} \big\{ \underset{j \in \{1,2\ldots t\}}{\min} \{d(E_i, E_j)\} \big
  \}
  \label{eq:greedy_selection}
\end{equation}

\subsection{Annotation selection}
\label{sub:annotation_selection}

Given a pool of annotation types $A = \{a_1, a_2, \ldots, a_L\}$, the goal of this step is to choose the most suitable type \anntype for \selframe. Following~\cite{active_image_seg, ksenia}, we formulate this problem as a Markov Decision Process and train a model using reinforcement learning (RL). The model observes the image of \selframe and its predicted mask $M_{f_*}^t$ and predicts the most suitable annotation type \anntype. This annotation type is then utilized by the annotator to generate a new mask $M_{f_*}^{t+1}$ for \selframe. 
The annotation is performed iteratively (e.g. 3 clicks are performed one by one). Therefore, we denote the annotation iteration as $g$. The input $M_{f_*}^t$ has an initial segmentation quality $\text{SQ}_1$ ($g=1$). 

\myparb{Environment.}
To give our model the ability to play the annotation selection game, the environment consists of the Human annotation and mask prediction stage (Sec.~\ref{sub:mask_pred}). 
The state of the environment consists of \selframe and its mask. Each step $g$ yields $\text{SQ}_g$ using the input action, which represents an annotation type from $A$.

\myparb{Reward.}
The reward function reflects the trade-off between the quality of $M_\text{\selframe}^t$ and the cost of the annotation type. Each $a \in A$ requires a different annotation cost denoted by $\theta_a$. The reward at $g$ is formulated by comparing SQ before and after annotation, divided by the total cost $tc$ at $g$ which is the sum of the costs $\theta_a$ of all annotation types until $g$:
\begin{equation}
    r = \frac{\text{SQ}_{g+1}-\text{SQ}_{g}}{tc} \quad . 
    \label{eq:reward}
\end{equation}
This equation captures the improvement of SQ, normalized by the total cost, and our model is trained to maximize this improvement while minimizing the annotation cost.

\myparb{Architecture.}
The model has two image encoders~\cite{SAM, resnet} in parallel branches, one for the frame $f_i$ and one for the mask $M_{f_*}^t$. The extracted embeddings from each encoder are then concatenated and fed into two linear layers. The first layer has $L$ outputs (possible annotation types), while the second layer has one output for the RL value.

\myparb{Training.}
Following its success in several other tasks~\cite{gpt_4,nlp_ppo}, we use Proximal Policy Optimization~\cite{ppo} (PPO) to train our model.
At training, we use the simulated masks described in Sec.~\ref{sub:frame_selection}. 
At each iteration $t$, we use \selframe and its corresponding mask $M_\text{\selframe}^t$ to play the annotation selection game and train our agent.
We perform multiple environment steps and the process terminates when we reach the maximum steps or the type of drawing a mask is selected.

\begin{figure*}[t]
  \centering
    \begin{subfigure}{0.33\linewidth}
         \centering
         \includegraphics[width=\linewidth]{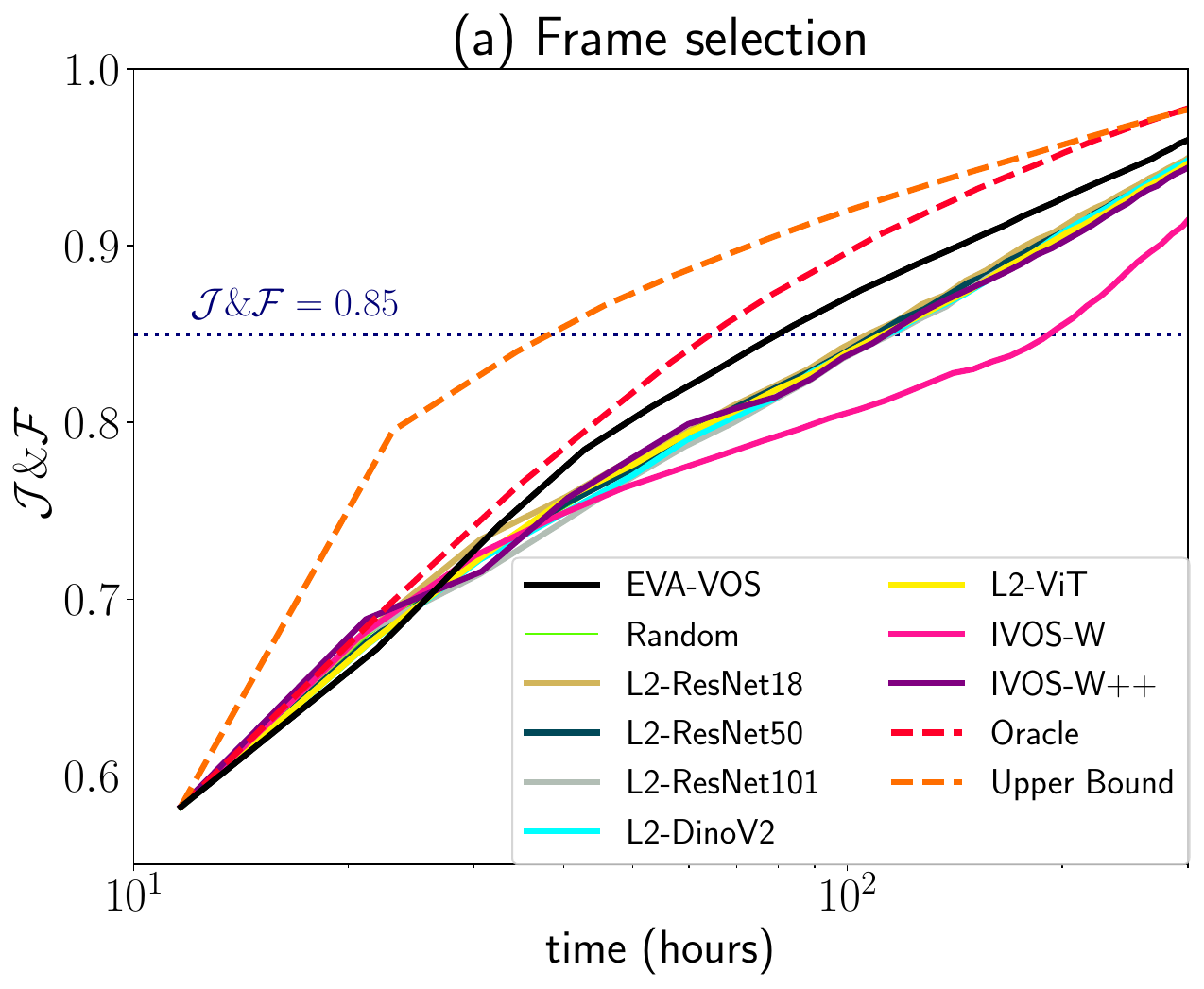}
    \end{subfigure}
    \begin{subfigure}{0.33\linewidth}
         \centering
         \includegraphics[width=\linewidth]{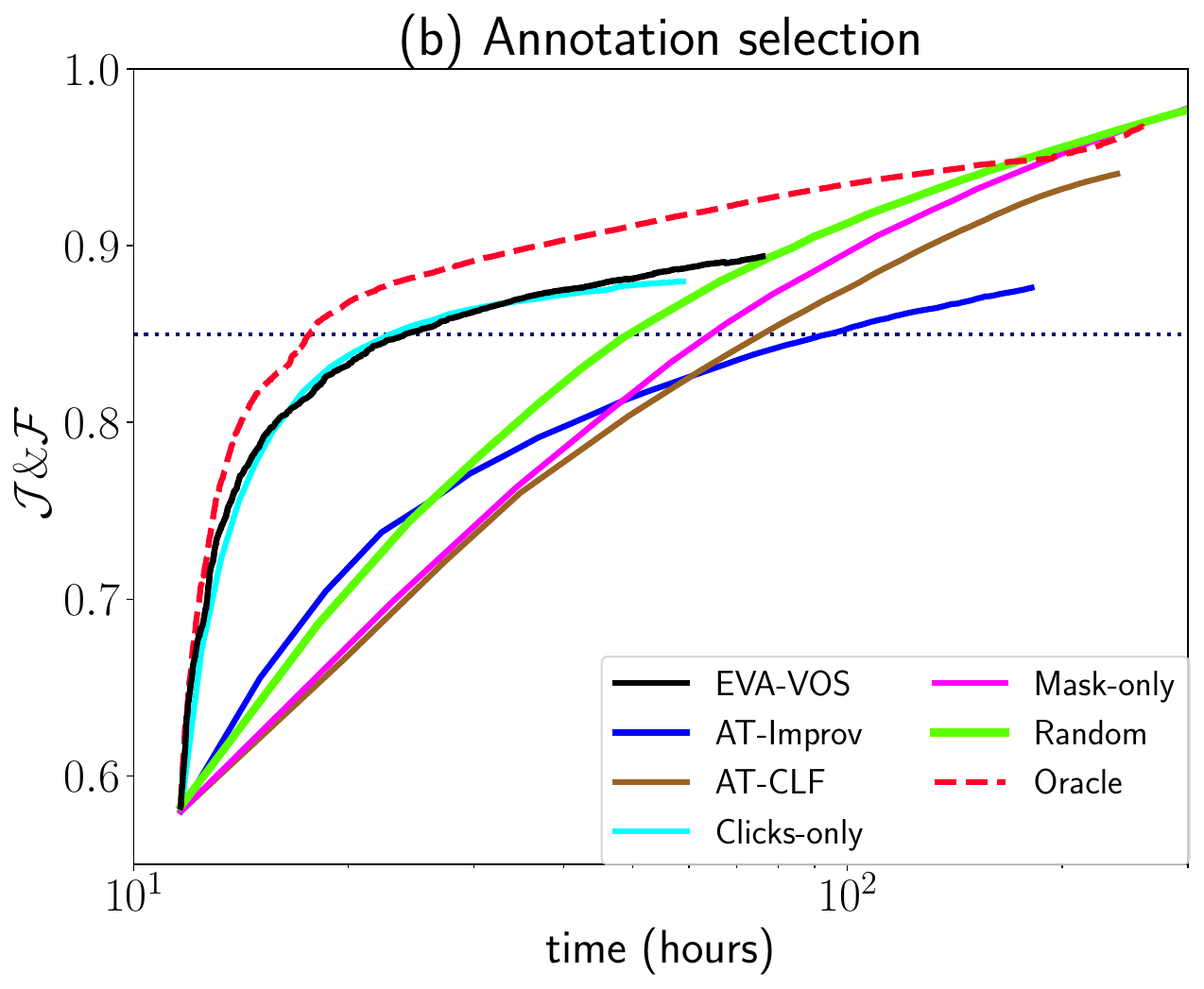}
     \end{subfigure}
     \begin{subfigure}{0.33\linewidth}
         \centering
         \includegraphics[width=\linewidth]{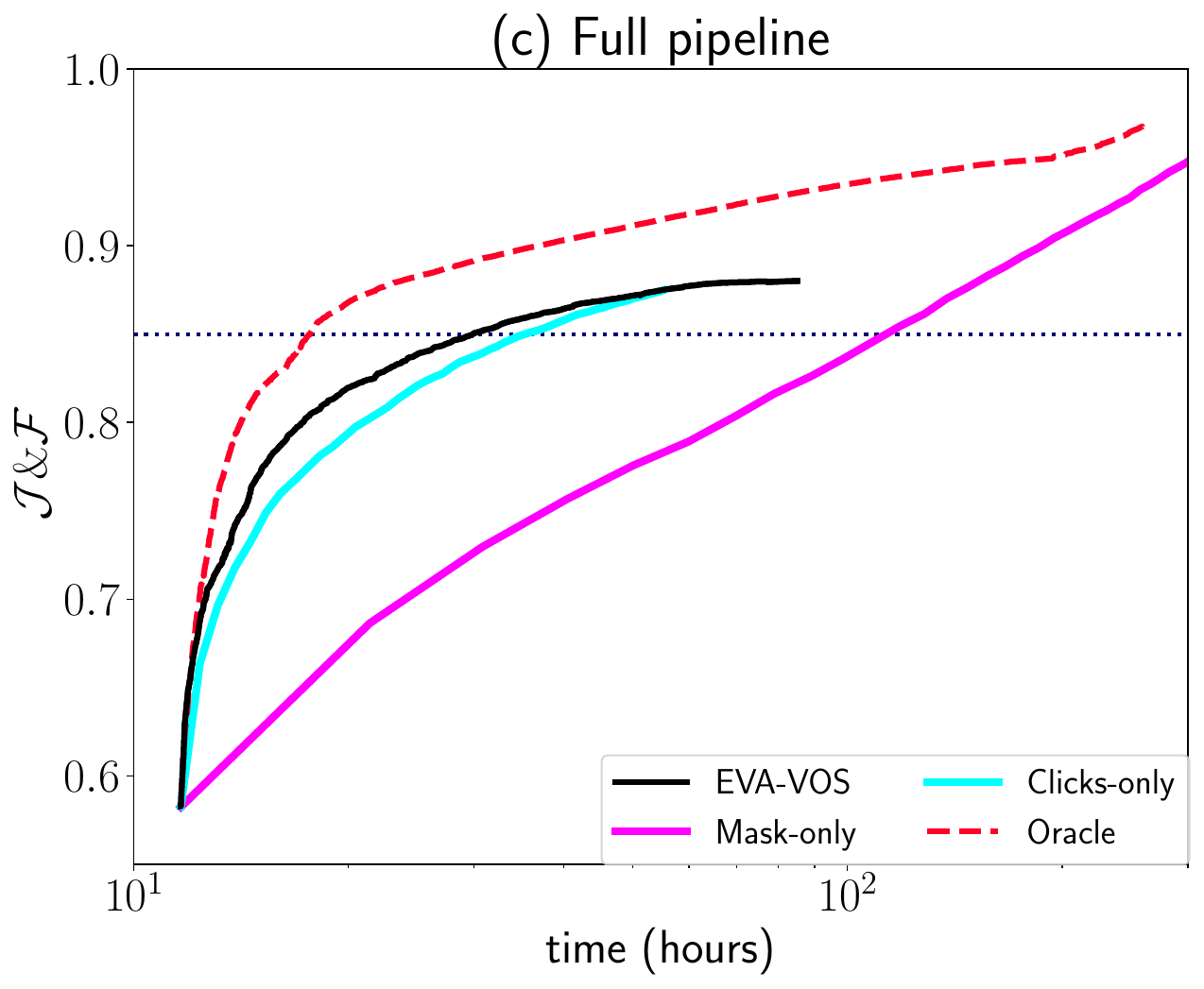}
     \end{subfigure}
   \caption{\small \textbf{Experimental results on MOSE.} We report the $\mathcal{J}\&\mathcal{F}$ accuracy as a function of annotation time in hours. (a) The effect of the frame selection stage (for fair comparison we use the same annotation type for all approaches). (b) The effect of the annotation selection stage using the same frame selection (oracle) for all approaches. (c) The results of our full pipeline. 
   }
   \label{fig:result_figs}
\end{figure*}

\myparb{Video Ranking.}
When \mname is used to annotate a collection of videos, similar to active learning, we use the predicted value $pv$ of our RL agent to estimate the improvement of each annotation at each video. This allows us to rank the videos and perform more annotation iterations in videos where the RL value is higher. To this end, for each iteration $t$, we first calculate the score $s$ of each video and perform an annotation iteration to the video with the maximum $s$. The score for each video is defined as a function of $pv$, the annotation iteration $t$, and annotation cost $\theta_a$:
\begin{equation}
    s = \frac{pv\cdot \gamma^t}{\theta_a} + c \quad , 
    \label{eq:ranking_score}
\end{equation}
where $\gamma<1$ and it allows us to prioritize videos with less annotations (smaller $t$). $\gamma$ is necessary since our agent has no information about the number of annotations of each video. Finally, $c$ scales the score $s$ into a positive range.


\subsection{Human annotation and mask prediction} 
\label{sub:mask_pred}
Here, the annotator interacts with the selected \selframe to create the input annotation type \anntype.
%
When \anntype is \emph{`mask drawing'}, the annotator draws a detailed $M_\text{\selframe}^{t+1}$.
Otherwise, i.e. clicks, this step predicts a new mask $M_\text{\selframe}^{t+1}$ using a pre-trained annotation-to-mask (A2M) model~\cite{SAM}. A2M predicts $M_\text{\selframe}^{t+1}$ based on the input weak annotation.

\myparb{A2M} predicts the new mask $M_\text{\selframe}^{t+1}$ of the selected frame \selframe with the input annotation type \anntype. There are various models that predict a segmentation mask given a weak annotation type~\cite{bearman16eccv,benenson19cvpr,khoreva2017simple,SAM,lin16cvpr}. 
Since our method is independent of this model, we opt for the recently introduced Segment Anything Model (SAM)~\cite{SAM}.
SAM can take as input the following annotation types: $A=\{$clicks (positive and negative), bounding boxes, masks, and text$\}$ to predict $M_\text{\selframe}^{t+1}$. When \anntype is anything but a number of clicks, SAM takes as input the annotation type \anntype and the current mask $M_{\text{\selframe}}^{t}$ of \selframe to predict $M_{\text{\selframe}}^{t+1}$. Otherwise, when \anntype is a number of clicks, SAM predicts $M_{\text{\selframe}}^{t+1}$ recursively. In particular, at each iteration, SAM takes as input one click and its own previous mask prediction to output a new segmentation mask. This process is repeated as many times as the number of clicks and the final prediction is the new mask $M_{\text{\selframe}}^{t+1}$ of the selected frame \selframe. It should be noted that this process is initialized with $M_{\text{\selframe}}^{t}$ which is predicted by the VOS~\cite{mivos} module and SAM can only take as input a mask from its own previous prediction. Therefore, we simulate clicks extracted by $M_{\text{\selframe}}^{t}$ and input them into SAM to generate a similar mask to the predicted by the VOS module.

\section{Experimental setting}
\label{sec:setting}
\myparb{Datasets.}
\textbf{DAVIS~17} contains 60 train and 30 validation videos. It provides high-quality annotated masks for each frame. 
 \textbf{MOSE} inherits videos from OVIS~\cite{ovis} and it is one of the largest available VOS dataset 
with 2149 videos, out of which only 1507 come with available ground-truth masks. In our work, to model long-range interactions we only consider videos with 15 to 104 frames leading to \dataset dataset with 1166 videos. We split it into 800 training, 150 validation, and 216 test videos.  

In our experiments, \mname is pre-trained on ImageNet~\cite{imagenet} and trained on \dataset, unless stated otherwise. Following the trend of zero-shot testing~\cite{clip}, to examine cross-dataset generalization, we evaluate \mname on the \dataset test set and on the DAVIS validation set.

\myparb{Metrics.}
To measure the segmentation quality of the predicted masks, we use both the intersection-over-union $\mathcal{J}$ and the contour accuracy $\mathcal{F}$~\cite{davis_16}. For this, we follow~\cite{davis_18} and use the curve of $\mathcal{J}\&\mathcal{F}$ vs time. 
We also report the annotation time in hours at $\mathcal{J}\&\mathcal{F}=\{0.75,0.80,0.85\}$ (different levels of human annotation agreement for instance segmentation~\cite{benenson19cvpr,gupta19cvpr,kirillov19cvpr,zhou17cvpr}), and the average \jandf up to 200 hours of annotation time. We consider 80 sec for drawing an object mask~\cite{coco} and 1.5 sec for each click plus 1 sec of overhead for the annotator to locate the object~\cite{bearman16eccv, benenson19cvpr, papadopoulos17cvpr}.

\myparb{Implementation details.}
\qnet consists of two ResNet-18~\cite{resnet}. We train it using SGD with $lr{=}10^{-5}$, batch size  64, 30 epochs, with $B{=}20$. The frame branch of the RL agent is the image encoder of SAM~\cite{SAM} while the mask branch is a ResNet-18~\cite{resnet}. The RL agent is trained using Adam~\cite{adam} and $lr{=}10^{-5}$ for 50K iterations.
For the video ranking, we estimate the hyper-parameters of Eq.~\eqref{eq:ranking_score} 
in the \dataset validation set. For all experiments, we use $\gamma$ to $0.7$ and $c$ to $-0.04$.
For the VOS module, we use a modified version of MiVOS~\cite{mivos}, where we discard the original interaction module~\cite{mivos} (as it only works with scribbles) and replace the propagation module with STCN~\cite{stcn} for faster propagation. %
To better examine the effect of frame and annotation selections, we pre-train this modified version only on  YouTubeVOS~\cite{youtube_vos} and not on DAVIS~\cite{davis_17}.
In our experiments, we consider two annotation types: \emph{`mask drawing'} and \emph{`corrective clicks'}~\cite{benenson19cvpr}. For \emph{`corrective clicks'}, the annotator clicks 3 times to improve the given segmentation and determines the number of positive and negative clicks. 
For simplicity, for the remainder of this work, we denote \emph{`mask drawing'} as Mask and \emph{`corrective clicks'} as Clicks.

\myparb{Human annotator simulation.}
In this work, we only perform experiments by simulating the human intervention.
Given $M_{\text{\selframe}}^t$ and the ground-truth mask of \selframe $m_g$, we simulate positive and negative clicks to prompt SAM~\cite{SAM} similar to how a human would. Initially, we identify all false-negative and false-positive pixels between $m_g$ and $M_{\text{\selframe}}^t$. Then, we determine the connected components of each error region, and the center of the largest component is selected as the click location, whether positive or negative.

\section{Experimental results}
\label{sec:experiments}
\begin{figure}[t]
  \centering
  \includegraphics[width=0.85\linewidth]{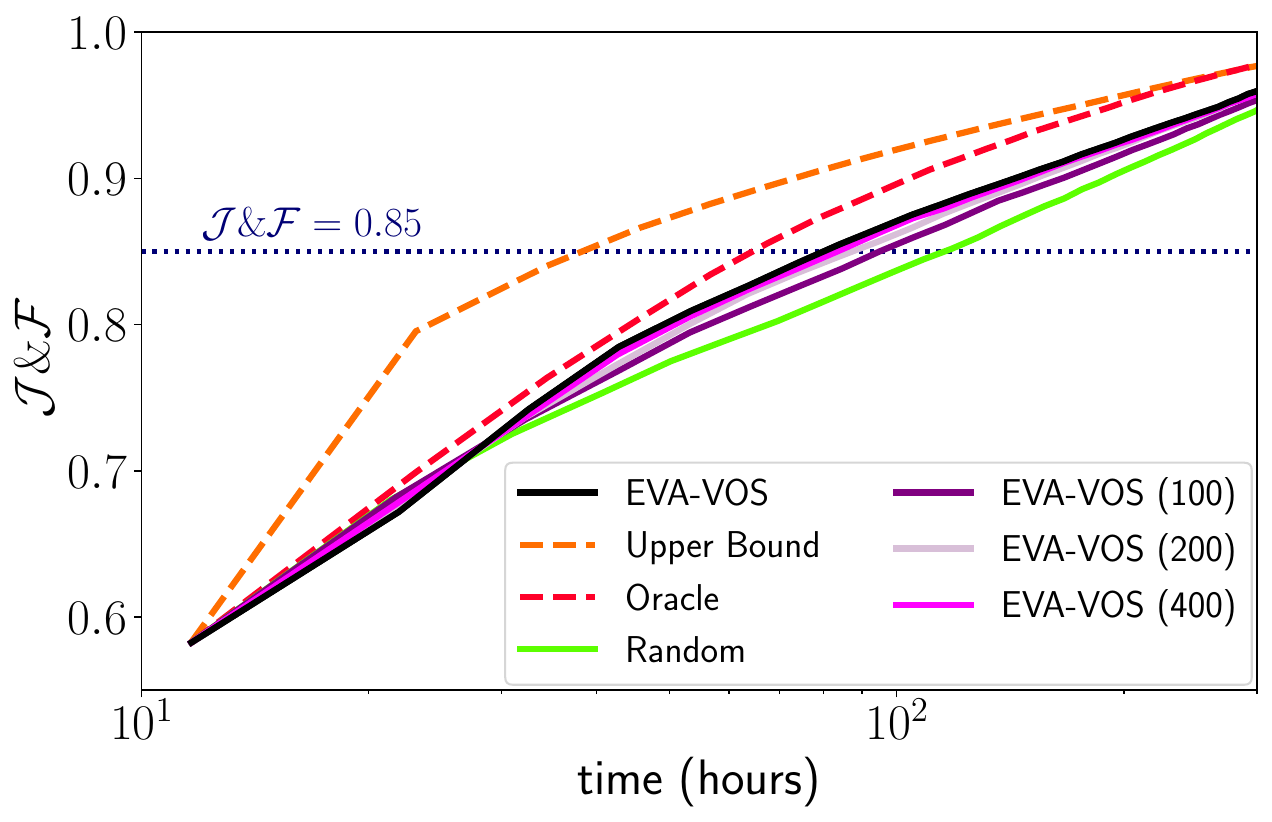}
   \caption{\textbf{Ablation study} on the number of training videos in the frame selection stage. 
   }
   \label{fig:abl_training_data}
\end{figure}

%
In all cases, we use figures for evaluation and show the segmentation quality (\jandf) of the predicted masks for all methods in relation to the annotation time on a log axis.  
 We first evaluate the frame selection and annotation selection (Sec.~\ref{sub:annotation_selection_experiments}) individually and show the results in Fig.~\ref{fig:result_figs}(a) and (b), respectively. Then, we analyze the results of our full pipeline (Sec.~\ref{sub:method_experiments}) and report results in Fig.~\ref{fig:result_figs}(c). 
 Finally, in Sec.~\ref{sub:gen_ability_exps} we examine the generalization ability of \mname, by performing a cross-dataset examination (Fig.~\ref{fig:ovis_davis_results}). 
 \begin{table}[t]
    \centering
    \caption{\textbf{Comparison of annotation methods on MOSE~\cite{mose}.} We report the human annotation time in hours for each method to reach different $\mathcal{J}\&\mathcal{F}$ values (0.75, 0.8, 0.85). We also report the average \jandf up to $200$ hours. At the top of the table, we report the oracle performance of oracle approaches for frame and/or annotation selection.
    \textbf{Bold} is the overall best-performing model, while \underline{Underline} is the best-performing frame selection approach that uses Mask-only as an annotation type.}
    \label{tab:1}
    \small
    \setlength{\tabcolsep}{2pt} 
   \begin{tabular}{@{}ll|ccc|c@{}}
        \toprule
        Annotation &
        Frame &
        \multicolumn{3}{c|}{Hours at \jandf$=$} & 
        \multirow{2}{*}{\rotatebox[origin=c]{0}{Avg \jandf $\uparrow$}} \\
        Selection & Selection & 0.75 & 0.80 & 0.85 $\downarrow$ & \\
        \midrule
        Mask-only&Oracle$^\star$          & 34.42 & 45.62 & 67.64  & 0.83 \\
        Clicks-only&Oracle$^\star$    & 14.05 & 15.65 & 22.85 & 0.87 \\
        Oracle$^\star$ & Oracle$^\star$  & 12.96 & 14.13 & 17.63 & 0.92 \\
        \midrule
        Mask-only&IVOS-W~\cite{IVOS-W}   & 39.37 & 94.33 & 192.26 & 0.78 \\
        Mask-only&IVOS-W++               & 40.53 & 59.81 & 113.93 & 0.79 \\
        Mask-only&L2-ResNet50            & 40.55 & 59.92 & 109.42 & 0.80 \\
        Mask-only&Random                 & 40.55 & 69.60 & 107.40 & 0.80 \\
        Mask-only&\underline{\mname}     & \underline{32.55} & \underline{53.26} & \underline{80.71} & \underline{0.82} \\
        \midrule
        Random&Random           & 24.08 & 36.10 & 65.84 & 0.85 \\
        Clicks-only&Random           & 15.32 & 21.22 & 35.10 & 0.86 \\
        \textbf{\mname}&\textbf{\mname} & \textbf{14.24} & \textbf{17.25} & \textbf{29.80} & \textbf{0.87} \\
        \bottomrule
    \end{tabular}
\end{table}

\subsection{Frame selection evaluation}
\label{sub:frame_selection_experiments}
Here, we evaluate the frame selection  (Sec.~\ref{sub:frame_selection}) and display the results in Fig.~\ref{fig:result_figs}(a). For a fair comparison among all methods, we use only Mask as an annotation type.

    

\begin{figure*}
  \centering
    \includegraphics[scale=0.465]{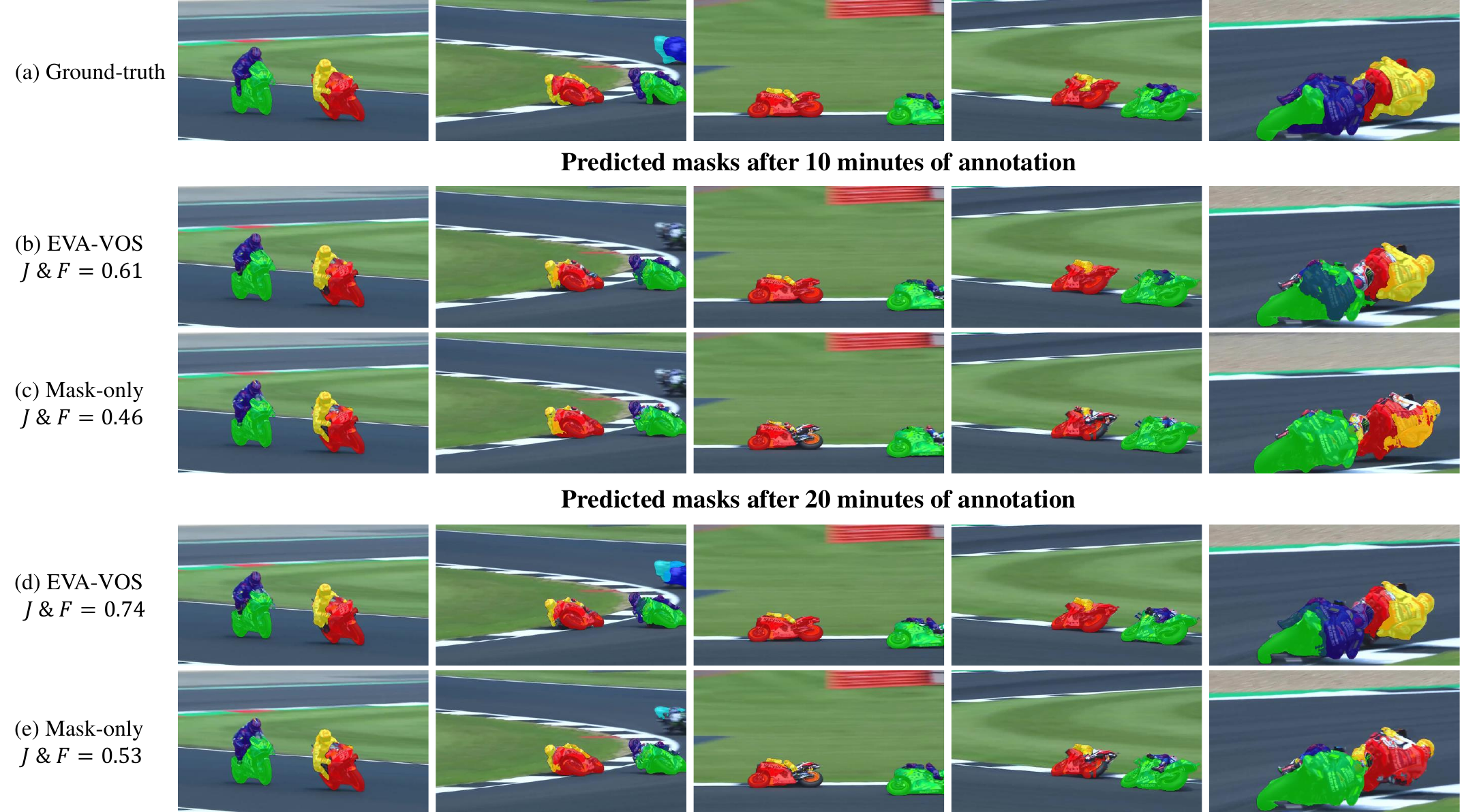}
    \caption{We compare \mname with Mask-only which resembles the traditional VOS annotation pipeline in one video from MOSE~\cite{mose}. The row (a) shows the ground-truth masks on 5 frames of a video with six annotated objects (purple, green, yellow, red, blue, and cyan). The rows (b) and (c) show the predicted masks of \mname and Mask-only after 10 minutes of annotation while the rows (d) and (e) after 20 minutes. We observe that \mname consistently outperforms Mask-only at different annotation budgets. For example, 
    in the second frame, \mname correctly segments the blue and cyan objects at 20 minutes, while Mask-only fails. Similarly, 
    in the third and fourth frames, \mname correctly segments the red object at 10 minutes, while Mask-only fails even after 20 minutes.}
    \label{fig:qualitative_results}
\end{figure*}

\myparb{Compared methods.}
We compare our method to a random baseline that selects frames randomly.
We also compare \mname to IVOS-W~\cite{IVOS-W}, which is the state-of-the-art frame selection method for VOS. Note that IVOS-W was originally designed to work only under a scribble-based iVOS scenario. Therefore, we modify it here to work for different annotation types, and for a fair comparison, we train this modified version in \dataset.
Furthermore, we implement and compare to it the IVOS-W++ in which we replace the RNN with a transformer~\cite{attention_is_all_you_need} and the double Q-Learning~\cite{double_dqn} with PPO~\cite{ppo}.
We also use powerful image encoders~\cite{resnet, vit, dinov2} pre-trained for image classification~\cite{russakovsky15ijcv} to compute embeddings in Eq.~\eqref{eq:greedy_selection} 
and we compare the results with \qnet.
We implement an oracle approach that selects the frame with the worst $\mathcal{J}\&\mathcal{F}$ and an upper bound approach that selects the frame that has the highest impact in the propagation stage after annotating it.

\mypar{Comparison to the state of the art.} 
Fig.~\ref{fig:result_figs}(a) shows that most methods have a similar performance close to Random. Instead, \mname consistently stands out and yields higher \jandf and in some cases almost identical to Oracle. We now analyze the results of all methods:
\\ \textit{Random} is shown as the green line in Fig.~\ref{fig:result_figs}(a). We run all random baselines 15 times and report the average result. 
\\ \textit{\mname (Ours)} is shown as the black line in Fig.~\ref{fig:result_figs}(a). Given the same annotation time, our framework consistently outperforms Random. For instance, we achieve $\mathcal{J}\&\mathcal{F}{=}0.85$ at 80.7 hours, 26.7 hours faster than Random.
\\ \textit{State-of-the-art} frame selection (IVOS-W~\cite{IVOS-W}) performs significantly worse than our method. We observe that our method reaches $\mathcal{J}\&\mathcal{F}$ of 0.85 2.3$\times$ faster than IVOS-W. 
\\ \textit{IVOS-W++} performs better than IVOS-W but our method achieves $\mathcal{J}\&\mathcal{F}$ of $0.85$ 1.4$\times$ faster.
\\ \textit{L2-Encoders} yield approximately the same performance as random. This shows that our task-specific \qnet learns much better representations and outperforms all pre-trained encoders that have even 28$\times$ more parameters.
\\ \textit{Oracle} is shown as the red dashed line in Fig.~\ref{fig:result_figs}(a). Interestingly, we observe that for low budgets (up to 40 hours), our method yields almost identical $\mathcal{J}\&\mathcal{F}$.
\\ \textit{Upper Bound} consistently outperforms the oracle indicating that the frame with the worst $\mathcal{J}\&\mathcal{F}$ is not the most impactful one. Interestingly, the upper bound is only 2.1$\times$ faster than our method at $\mathcal{J}\&\mathcal{F}=0.85$.

\myparb{Ablation study on the number of training videos.}
QNet is trained on \dataset (800 videos). We retrain it using 100, 200, and 400 videos to examine the impact of the training data. In Fig.~\ref{fig:abl_training_data}, we compare all of our frame selection models with the random method as it performs approximately the same as the L2-Encoders and IVOS-W++. We observe that all of our models outperform Random and even with a reduced training dataset, i.e, 400 videos, \qnet showcases $\mathcal{J}\&\mathcal{F}$ that closely aligns with the initial model trained with a larger dataset of 800 videos. This reveals the robustness and effectiveness of our training process (Sec.~\ref{sub:frame_selection}). 


\subsection{Annotation selection evaluation}
\label{sub:annotation_selection_experiments}
Here, we evaluate only our annotation selection stage. For a fair comparison, we set the frame selection for all approaches to oracle, i.e., the frame with the worst $\mathcal{J}\&\mathcal{F}$ is selected to be annotated at each iteration (results in Fig.~\ref{fig:result_figs}(b)).

\mypar{Compared methods.}
To examine the design choice of RL (Sec.~\ref{sub:annotation_selection}), we implement two alternatives for annotation selection. The first one is AT-Improv (Annotation Type Improvement), which is trained to regress the improvement of each available annotation type. It selects the annotation type that maximizes Eq.~\eqref{eq:reward}. The second one is AT-CLF (Annotation Type Classification), and it is trained to classify each \selframe into an annotation type.
Furthermore, we compare against approaches that consider only one annotation type (Clicks or Mask). We also compare against a random baseline that selects \anntype randomly, and an oracle approach that selects \anntype using Eq.\eqref{eq:reward}, i.e., it selects 
the \anntype that yields the maximum quality improvement normalized by the annotation cost. Moreover, the oracle approach ranks the videos using Eq.~\eqref{eq:reward}, while our approach uses Eq.~\eqref{eq:ranking_score}.

\mypar{Comparison to annotation selection methods.} In Fig.~\ref{fig:result_figs}(b) we observe the impact of annotation selection since Clicks-only plateaus at lower \jandf while Mask-only is significantly slower. Furthermore, methods that do not select the annotation type wisely, perform worse than Mask-only. We now analyze the results of all methods:\\
\textit{EVA-VOS (Ours)} is shown as the black line in Fig.~\ref{fig:result_figs}(b). It reaches $\mathcal{J}\&\mathcal{F}=0.85$ in only 29.8 hours.
\\
\textit{AT-Improv, AT-CLF} (blue and brown lines in Fig.~\ref{fig:result_figs}(b)) perform significantly worse than \mname which is trained using RL instead of supervised learning.
\\
\textit{Random} is shown as the green line in Fig.~\ref{fig:result_figs}(b). Even though it reaches $\mathcal{J}\&\mathcal{F}=0.9$ at a similar time as our method, it performs significantly worse at lower budgets (e.g., we yield $\mathcal{J}\&\mathcal{F}=0.85$ 1.9$\times$ faster). 
%
\textit{Mask-only} performs consistently worse than random at all budgets, indicating that the traditional way of manually drawing object mask~\cite{davis_16,ovis,youtube_vos} is not a good approach.
\\
\textit{Clicks-only} performs on par with our method at low annotation budgets. However, it plateaus quickly at lower $\mathcal{J}\&\mathcal{F}$ values and it is not able to reach $\mathcal{J}\&\mathcal{F}=0.9$, whereas our method can yield higher $\mathcal{J}\&\mathcal{F}$ at larger budgets.
\\
\textit{Oracle} (red dashed line in Fig.~\ref{fig:result_figs}(b)) performs on par with our method at low budgets. Oracle performs better in very high annotation budgets that reach high $\mathcal{J}\&\mathcal{F}$ above $0.85$. 

\begin{figure}[t]
    \centering
    \includegraphics[width=0.85\linewidth]{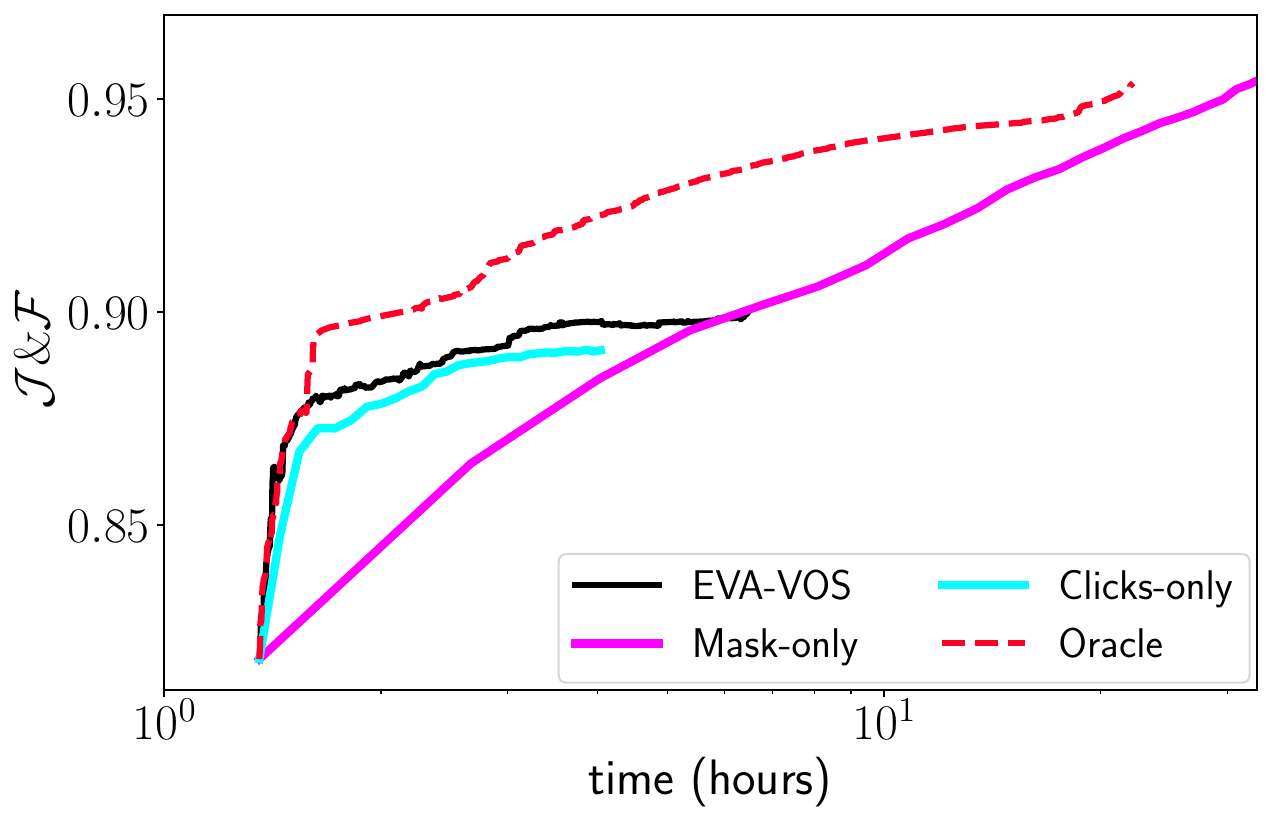}
    \caption{\textbf{\mname results on DAVIS 17.} We report the $\mathcal{J}\&\mathcal{F}$ accuracy as a function of annotation time in hours at log scale.}
   \label{fig:ovis_davis_results}
\end{figure}

\subsection{Frame and Annotation selection evaluation}
\label{sub:method_experiments}
We evaluate here our full pipeline, showing the effect of both selection modules (Fig.~\ref{fig:result_figs}(c) and Tab.~\ref{tab:1}).

\mypar{Compared methods.}
Similar to Sec.~\ref{sub:annotation_selection_experiments}, we compare our method to Clicks-only and Masks-only which select a random frame and consider only one annotation type. Additionally, we compare against Oracle, which uses both oracle frame selection and annotation selection.

\mypar{Comparison of annotation methods.}\
In Fig.~\ref{fig:result_figs}(c), we examine the impact of both selection modules. Overall, we observe that a large amount of annotation time can be saved when annotating with more than one type and wisely selecting frames. We now present the results of all methods:\\
\textit{EVA-VOS (Ours)} is shown as the black line in Fig.~\ref{fig:result_figs}(c) and yield a $\mathcal{J}\&\mathcal{F}$ of $0.85$ in 29.8 hours (also, last row in Tab.~\ref{tab:1}).
\\
\textit{Oracle} uses both oracle frame selection and annotation selection and shows the trade-off that \mname could achieve with an ideal oracle training scenario.
\\
\textit{Mask-Only} resembles the traditional way of annotating videos with segmentation masks~\cite{youtube_vos, uvo, visor, ovis, mose}. Our method performs significantly better and achieves a 3.5$\times$ speed up compared to Mask-only at $\mathcal{J}\&\mathcal{F}=0.85$ (Tab.~\ref{tab:1}).
\\
\textit{Click-Only} performs similarly to \mname at 50 hours but has a worse trade-off for either lower or higher budgets.



Tab.~\ref{tab:1} compares \mname with the best frame and annotation selection approaches presented in Fig.~\ref{fig:result_figs}. We quantify performance using the human annotation time in hours for each method to reach different \jandf values and the average \jandf up to 200 hours. Similar to Fig.~\ref{fig:result_figs}, we observe that \mname overall outperforms all approaches, thus supporting our hypothesis that selecting frames and annotation type leads to both performance and time gains.

\mypar{Qualitative results.} 
In Fig.~\ref{fig:qualitative_results}, we qualitatively compare \mname to Mask-only. Specifically, we illustrate the predicted masks of each method at different annotation budgets. We observe that \mname predicts more accurate masks faster than Mask-only.

\subsection{\mname generalization ability}
\label{sub:gen_ability_exps}

We now evaluate \mname in DAVIS~17~\cite{davis_17} without training any of our components on it. 
Similar to Sec.~\ref{sub:method_experiments}, we compare our method to Clicks-only and Masks-only. Fig.~\ref{fig:ovis_davis_results} illustrates the results, where we observe that \mname performs on par with Clicks-only and significantly outperforms Masks-only in lower annotation budgets.


\section{Conclusions}
\label{sec:conclusions}
We presented an alternative and efficient way to annotate objects in videos with segmentation masks. 
Our \mname framework shows significant gains in terms of annotation time (3.5$\times$ speed up) compared to the traditional, manual way of annotating objects in videos.
Our experiments, especially on the challenging MOSE dataset, show that our framework reduces the total human annotation time while leading to high-quality segmentation masks for the videos. 

\mypar{Acknowledgements.}
D. Papadopoulos was supported by the DFF Sapere Aude Starting Grant ``ACHILLES''. V. Kalogeiton was supported by a Hi! PARIS grant and the ANR-22-CE23-0007. We would like to thank P. Pegios, J. Parslov, E. Riise, and Y. Benigmim for proofreading.

{\small
\bibliographystyle{ieee_fullname}
\bibliography{references}

\begin{thebibliography}{10}\itemsep=-1pt

\bibitem{acuna18cvpr}
David Acuna, Huan Ling, Amlan Kar, and Sanja Fidler.
\newblock Efficient interactive annotation of segmentation datasets with polygon-rnn++.
\newblock In {\em CVPR}, 2018.

\bibitem{prop_4}
Linchao Bao, Baoyuan Wu, and Wei Liu.
\newblock Cnn in mrf: Video object segmentation via inference in a cnn-based higher-order spatio-temporal mrf.
\newblock In {\em CVPR}, 2018.

\bibitem{v_editing_2}
Omer Bar-Tal, Dolev Ofri-Amar, Rafail Fridman, Yoni Kasten, and Tali Dekel.
\newblock Text2live: Text-driven layered image and video editing.
\newblock In {\em NeurIPS}, 2022.

\bibitem{bearman16eccv}
A. Bearman, O. Russakovsky, V. Ferrari, and L. Fei-Fei.
\newblock What's the point: Semantic segmentation with point supervision.
\newblock In {\em ECCV}, 2016.

\bibitem{xmem_plus}
Maksym Bekuzarov, Ariana Bermudez, Joon-Young Lee, and Hao Li.
\newblock Xmem++: Production-level video segmentation from few annotated frames.
\newblock In {\em ICCV}, 2023.

\bibitem{benenson19cvpr}
Rodrigo Benenson, Stefan Popov, and Vittorio Ferrari.
\newblock Large-scale interactive object segmentation with human annotators.
\newblock In {\em CVPR}, 2019.

\bibitem{on_1}
Sergi Caelles, Kevis-Kokitsi Maninis, Jordi Pont-Tuset, Laura Leal-Taix{\'e}, Daniel Cremers, and Luc Van~Gool.
\newblock One-shot video object segmentation.
\newblock In {\em CVPR}, 2017.

\bibitem{davis_18}
Sergi Caelles, Alberto Montes, Kevis-Kokitsi Maninis, Yuhua Chen, Luc {Van Gool}, Federico Perazzi, and Jordi Pont-Tuset.
\newblock The 2018 davis challenge on video object segmentation.
\newblock {\em arXiv:1803.00557}, 2018.

\bibitem{active_image_seg}
Arantxa Casanova, Pedro~O. Pinheiro, Negar Rostamzadeh, and Christopher~J. Pal.
\newblock Reinforced active learning for image segmentation.
\newblock In {\em ICLR}, 2020.

\bibitem{castrejon17cvpr}
L. Castrejon, K. Kundu, R. Urtasun, and S. Fidler.
\newblock Annotating object instances with a {Polygon-RNN}.
\newblock In {\em CVPR}, 2017.

\bibitem{chen2020scribblebox}
Bowen Chen, Huan Ling, Xiaohui Zeng, Jun Gao, Ziyue Xu, and Sanja Fidler.
\newblock Scribblebox: Interactive annotation framework for video object segmentation.
\newblock In {\em ECCV}, 2020.

\bibitem{prop_6}
Xi Chen, Zuoxin Li, Ye Yuan, Gang Yu, Jianxin Shen, and Donglian Qi.
\newblock State-aware tracker for real-time video object segmentation.
\newblock In {\em CVPR}, 2020.

\bibitem{chen2018blazingly}
Yuhua Chen, Jordi Pont-Tuset, Alberto Montes, and Luc Van~Gool.
\newblock Blazingly fast video object segmentation with pixel-wise metric learning.
\newblock In {\em CVPR}, 2018.

\bibitem{xmem}
Ho~Kei Cheng and Alexander~G Schwing.
\newblock Xmem: Long-term video object segmentation with an atkinson-shiffrin memory model.
\newblock In {\em ECCV}, 2022.

\bibitem{mivos}
Ho~Kei Cheng, Yu-Wing Tai, and Chi-Keung Tang.
\newblock Modular interactive video object segmentation: Interaction-to-mask, propagation and difference-aware fusion.
\newblock In {\em CVPR}, 2021.

\bibitem{stcn}
Ho~Kei Cheng, Yu-Wing Tai, and Chi-Keung Tang.
\newblock Rethinking space-time networks with improved memory coverage for efficient video object segmentation.
\newblock In {\em NeurIPS}, 2021.

\bibitem{prop_5}
Jingchun Cheng, Yi-Hsuan Tsai, Wei-Chih Hung, Shengjin Wang, and Ming-Hsuan Yang.
\newblock Fast and accurate online video object segmentation via tracking parts.
\newblock In {\em CVPR}, 2018.

\bibitem{sam_track}
Yangming Cheng, Liulei Li, Yuanyou Xu, Xiaodi Li, Zongxin Yang, Wenguan Wang, and Yi Yang.
\newblock Segment and track anything.
\newblock {\em arXiv preprint arXiv:2305.06558}, 2023.

\bibitem{dai2021video}
Kenan Dai, Jie Zhao, Lijun Wang, Dong Wang, Jianhua Li, Huchuan Lu, Xuesheng Qian, and Xiaoyun Yang.
\newblock Video annotation for visual tracking via selection and refinement.
\newblock In {\em ICCV}, 2021.

\bibitem{visor}
Ahmad Darkhalil, Dandan Shan, Bin Zhu, Jian Ma, Amlan Kar, Richard Higgins, Sanja Fidler, David Fouhey, and Dima Damen.
\newblock Epic-kitchens visor benchmark: Video segmentations and object relations.
\newblock In {\em NeurIPS}, 2022.

\bibitem{imagenet}
Jia Deng, Wei Dong, Richard Socher, Li-Jia Li, Kai Li, and Li Fei-Fei.
\newblock Imagenet: A large-scale hierarchical image database.
\newblock In {\em CVPR}, 2009.

\bibitem{mose}
Henghui Ding, Chang Liu, Shuting He, Xudong Jiang, Philip~HS Torr, and Song Bai.
\newblock {MOSE}: A new dataset for video object segmentation in complex scenes.
\newblock In {\em ICCV}, 2023.

\bibitem{vit}
Alexey Dosovitskiy, Lucas Beyer, Alexander Kolesnikov, Dirk Weissenborn, Xiaohua Zhai, Thomas Unterthiner, Mostafa Dehghani, Matthias Minderer, Georg Heigold, Sylvain Gelly, Jakob Uszkoreit, and Neil Houlsby.
\newblock An image is worth 16x16 words: Transformers for image recognition at scale.
\newblock {\em ICLR}, 2021.

\bibitem{bubble_nets}
Brent~A. Griffin and Jason~J. Corso.
\newblock Bubblenets: Learning to select the guidance frame in video object segmentation by deep sorting frames.
\newblock In {\em CVPR}, 2019.

\bibitem{openvis}
Pinxue Guo, Tony Huang, Peiyang He, Xuefeng Liu, Tianjun Xiao, Zhaoyu Chen, and Wenqiang Zhang.
\newblock Openvis: Open-vocabulary video instance segmentation.
\newblock {\em arXiv preprint arXiv:2305.16835}, 2023.

\bibitem{gupta19cvpr}
Agrim Gupta, Piotr Dollar, and Ross Girshick.
\newblock Lvis: A dataset for large vocabulary instance segmentation.
\newblock In {\em CVPR}, 2019.

\bibitem{resnet}
Kaiming He, Xiangyu Zhang, Shaoqing Ren, and Jian Sun.
\newblock Deep residual learning for image recognition.
\newblock In {\em CVPR}, 2016.

\bibitem{GIS-RAmap}
Yuk Heo, Yeong~Jun Koh, and Chang-Su Kim.
\newblock Guided interactive video object segmentation using reliability-based attention maps.
\newblock In {\em CVPR}, 2021.

\bibitem{matching_2}
Yuan-Ting Hu, Jia-Bin Huang, and Alexander~G Schwing.
\newblock Videomatch: Matching based video object segmentation.
\newblock In {\em ECCV}, 2018.

\bibitem{prop_3}
Varun Jampani, Raghudeep Gadde, and Peter~V Gehler.
\newblock Video propagation networks.
\newblock In {\em CVPR}, 2017.

\bibitem{prop_2}
Won-Dong Jang and Chang-Su Kim.
\newblock Online video object segmentation via convolutional trident network.
\newblock In {\em CVPR}, 2017.

\bibitem{v_editing_1}
Yoni Kasten, Dolev Ofri, Oliver Wang, and Tali Dekel.
\newblock Layered neural atlases for consistent video editing.
\newblock 2021.

\bibitem{khoreva2017simple}
Anna Khoreva, Rodrigo Benenson, Jan Hosang, Matthias Hein, and Bernt Schiele.
\newblock Simple does it: Weakly supervised instance and semantic segmentation.
\newblock In {\em CVPR}, 2017.

\bibitem{davis_rvos}
Anna Khoreva, Anna Rohrbach, and Brent Schiele.
\newblock Video object segmentation with referring expressions.
\newblock In {\em ECCV Workshops}, pages 0--0, 2018.

\bibitem{adam}
Diederik~P. Kingma and Jimmy Ba.
\newblock Adam: A method for stochastic optimization.
\newblock In {\em ICLR}, 2017.

\bibitem{kirillov19cvpr}
Alexander Kirillov, Kaiming He, Ross Girshick, Carsten Rother, and Piotr Doll{\'a}r.
\newblock Panoptic segmentation.
\newblock In {\em CVPR}, 2019.

\bibitem{SAM}
Alexander Kirillov, Eric Mintun, Nikhila Ravi, Hanzi Mao, Chloe Rolland, Laura Gustafson, Tete Xiao, Spencer Whitehead, Alexander~C Berg, Wan-Yen Lo, et~al.
\newblock Segment anything.
\newblock {\em arXiv preprint arXiv:2304.02643}, 2023.

\bibitem{ksenia}
Ksenia Konyushkova, Raphael Sznitman, and Pascal Fua.
\newblock Learning active learning from data.
\newblock In {\em NeurIPS}, 2017.

\bibitem{polyp_sam}
Yuheng Li, Mingzhe Hu, and Xiaofeng Yang.
\newblock Polyp-sam: Transfer sam for polyp segmentation.
\newblock {\em arXiv preprint arXiv:2305.00293}, 2023.

\bibitem{li18cvpr}
Zhuwen Li, Qifeng Chen, and Vladlen Koltun.
\newblock Interactive image segmentation with latent diversity.
\newblock In {\em CVPR}, 2018.

\bibitem{lin16cvpr}
Di Lin, Jifeng Dai, Jiaya Jia, Kaiming He, and Jian Sun.
\newblock Scribblesup: Scribble-supervised convolutional networks for semantic segmentation.
\newblock In {\em CVPR}, 2016.

\bibitem{coco}
Tsung-Yi Lin, Michael Maire, Serge Belongie, James Hays, Pietro Perona, Deva Ramanan, Piotr Doll{\'a}r, and C~Lawrence Zitnick.
\newblock Microsoft coco: Common objects in context.
\newblock In {\em ECCV}, 2014.

\bibitem{matching_6_swem}
Zhihui Lin, Tianyu Yang, Maomao Li, Ziyu Wang, Chun Yuan, Wenhao Jiang, and Wei Liu.
\newblock Swem: Towards real-time video object segmentation with sequential weighted expectation-maximization.
\newblock In {\em CVPR}, 2022.

\bibitem{ling19cvpr}
Huan Ling, Jun Gao, Amlan Kar, Wenzheng Chen, and Sanja Fidler.
\newblock Fast interactive object annotation with curve-gcn.
\newblock In {\em CVPR}, 2019.

\bibitem{med_sam}
Jun Ma and Bo Wang.
\newblock Segment anything in medical images.
\newblock {\em arXiv preprint arXiv:2304.12306}, 2023.

\bibitem{on_5}
K-K Maninis, Sergi Caelles, Yuhua Chen, Jordi Pont-Tuset, Laura Leal-Taix{\'e}, Daniel Cremers, and Luc Van~Gool.
\newblock Video object segmentation without temporal information.
\newblock {\em PAMI}, 2018.

\bibitem{on_6}
Tim Meinhardt and Laura Leal-Taix{\'e}.
\newblock Make one-shot video object segmentation efficient again.
\newblock {\em NeurIPS}, 2020.

\bibitem{av_sam}
Shentong Mo and Yapeng Tian.
\newblock Av-sam: Segment anything model meets audio-visual localization and segmentation.
\newblock {\em arXiv preprint arXiv:2305.01836}, 2023.

\bibitem{matching_4_stm}
Seoung~Wug Oh, Joon-Young Lee, Ning Xu, and Seon~Joo Kim.
\newblock Video object segmentation using space-time memory networks.
\newblock In {\em ICCV}, 2019.

\bibitem{gpt_4}
OpenAI.
\newblock Gpt-4 technical report, 2023.

\bibitem{dinov2}
Maxime Oquab, Timothée Darcet, Theo Moutakanni, Huy~V. Vo, Marc Szafraniec, Vasil Khalidov, Pierre Fernandez, Daniel Haziza, Francisco Massa, Alaaeldin El-Nouby, Russell Howes, Po-Yao Huang, Hu Xu, Vasu Sharma, Shang-Wen Li, Wojciech Galuba, Mike Rabbat, Mido Assran, Nicolas Ballas, Gabriel Synnaeve, Ishan Misra, Herve Jegou, Julien Mairal, Patrick Labatut, Armand Joulin, and Piotr Bojanowski.
\newblock Dinov2: Learning robust visual features without supervision, 2023.

\bibitem{nlp_ppo}
Long Ouyang, Jeffrey Wu, Xu Jiang, Diogo Almeida, Carroll Wainwright, Pamela Mishkin, Chong Zhang, Sandhini Agarwal, Katarina Slama, Alex Ray, John Schulman, Jacob Hilton, Fraser Kelton, Luke Miller, Maddie Simens, Amanda Askell, Peter Welinder, Paul~F Christiano, Jan Leike, and Ryan Lowe.
\newblock Training language models to follow instructions with human feedback.
\newblock In {\em NeurIPS}, 2022.

\bibitem{papadopoulos17iccv}
Dim~P Papadopoulos, Jasper~RR Uijlings, Frank Keller, and Vittorio Ferrari.
\newblock Extreme clicking for efficient object annotation.
\newblock In {\em ICCV}, 2017.

\bibitem{papadopoulos17cvpr}
Dim~P Papadopoulos, Jasper~RR Uijlings, Frank Keller, and Vittorio Ferrari.
\newblock Training object class detectors with click supervision.
\newblock In {\em CVPR}, 2017.

\bibitem{papadopoulos21iccv}
Dim~P Papadopoulos, Ethan Weber, and Antonio Torralba.
\newblock Scaling up instance annotation via label propagation.
\newblock In {\em ICCV}, 2021.

\bibitem{prop_1}
Federico Perazzi, Anna Khoreva, Rodrigo Benenson, Bernt Schiele, and Alexander Sorkine-Hornung.
\newblock Learning video object segmentation from static images.
\newblock In {\em CVPR}, 2017.

\bibitem{davis_16}
Federico Perazzi, Jordi Pont-Tuset, Brian McWilliams, Luc Van~Gool, Markus Gross, and Alexander Sorkine-Hornung.
\newblock A benchmark dataset and evaluation methodology for video object segmentation.
\newblock In {\em CVPR}, 2016.

\bibitem{davis_17}
Jordi Pont-Tuset, Federico Perazzi, Sergi Caelles, Pablo Arbel{\'a}ez, Alex Sorkine-Hornung, and Luc Van~Gool.
\newblock The 2017 davis challenge on video object segmentation.
\newblock {\em arXiv preprint arXiv:1704.00675}, 2017.

\bibitem{ovis}
Jiyang Qi, Yan Gao, Yao Hu, Xinggang Wang, Xiaoyu Liu, Xiang Bai, Serge Belongie, Alan Yuille, Philip~HS Torr, and Song Bai.
\newblock Occluded video instance segmentation: A benchmark.
\newblock {\em IJCV}, 2022.

\bibitem{clip}
Alec Radford, Jong~Wook Kim, Chris Hallacy, Aditya Ramesh, Gabriel Goh, Sandhini Agarwal, Girish Sastry, Amanda Askell, Pamela Mishkin, Jack Clark, et~al.
\newblock Learning transferable visual models from natural language supervision.
\newblock In {\em PMLR}, 2021.

\bibitem{sam_pt}
Frano Rajič, Lei Ke, Yu-Wing Tai, Chi-Keung Tang, Martin Danelljan, and Fisher Yu.
\newblock Segment anything meets point tracking.
\newblock {\em arXiv:2307.01197}, 2023.

\bibitem{russakovsky15ijcv}
O. Russakovsky, J. Deng, H. Su, J. Krause, S. Satheesh, S. Ma, Z. Huang, A. Karpathy, A. Khosla, M. Bernstein, A. Berg, and L. Fei-Fei.
\newblock {ImageNet} large scale visual recognition challenge.
\newblock {\em IJCV}, 2015.

\bibitem{ppo}
John Schulman, Filip Wolski, Prafulla Dhariwal, Alec Radford, and Oleg Klimov.
\newblock Proximal policy optimization algorithms.
\newblock In {\em PMLR}, 2017.

\bibitem{urvos}
Seonguk Seo, Joon-Young Lee, and Bohyung Han.
\newblock Urvos: Unified referring video object segmentation network with a large-scale benchmark.
\newblock In {\em EECV}, 2020.

\bibitem{matching_1}
Jae Shin~Yoon, Francois Rameau, Junsik Kim, Seokju Lee, Seunghak Shin, and In So~Kweon.
\newblock Pixel-level matching for video object segmentation using convolutional neural networks.
\newblock In {\em ICCV}, 2017.

\bibitem{vost}
Pavel Tokmakov, Jie Li, and Adrien Gaidon.
\newblock Breaking the" object" in video object segmentation.
\newblock {\em arXiv preprint arXiv:2212.06200}, 2022.

\bibitem{double_dqn}
Hado van Hasselt, Arthur Guez, and David Silver.
\newblock Deep reinforcement learning with double q-learning.
\newblock {\em Proceedings of the AAAI Conference on Artificial Intelligence}, 2016.

\bibitem{attention_is_all_you_need}
Ashish Vaswani, Noam Shazeer, Niki Parmar, Jakob Uszkoreit, Llion Jones, Aidan~N Gomez, \L~ukasz Kaiser, and Illia Polosukhin.
\newblock Attention is all you need.
\newblock In {\em NeurIPS}, 2017.

\bibitem{prop_7}
Carles Ventura, Miriam Bellver, Andreu Girbau, Amaia Salvador, Ferran Marques, and Xavier Giro-i Nieto.
\newblock Rvos: End-to-end recurrent network for video object segmentation.
\newblock In {\em CVPR}, 2019.

\bibitem{matching_3}
Paul Voigtlaender, Yuning Chai, Florian Schroff, Hartwig Adam, Bastian Leibe, and Liang-Chieh Chen.
\newblock Feelvos: Fast end-to-end embedding learning for video object segmentation.
\newblock In {\em CVPR}, 2019.

\bibitem{on_2}
Paul Voigtlaender and Bastian Leibe.
\newblock Online adaptation of convolutional neural networks for video object segmentation.
\newblock {\em BMCV}, 2017.

\bibitem{clicks_2}
Stephane Vujasinovic, Sebastian Bullinger, Stefan Becker, Norbert Scherer-Negenborn, Michael Arens, and Rainer Stiefelhagen.
\newblock Revisiting click-based interactive video object segmentation.
\newblock In {\em ICIP}, 2022.

\bibitem{wang2019fast}
Qiang Wang, Li Zhang, Luca Bertinetto, Weiming Hu, and Philip~HS Torr.
\newblock Fast online object tracking and segmentation: A unifying approach.
\newblock In {\em CVPR}, 2019.

\bibitem{cap_anything}
Teng Wang, Jinrui Zhang, Junjie Fei, Yixiao Ge, Hao Zheng, Yunlong Tang, Zhe Li, Mingqi Gao, Shanshan Zhao, Ying Shan, et~al.
\newblock Caption anything: Interactive image description with diverse multimodal controls.
\newblock {\em arXiv preprint arXiv:2305.02677}, 2023.

\bibitem{vs_1}
Ting-Chun Wang, Ming-Yu Liu, Andrew Tao, Guilin Liu, Jan Kautz, and Bryan Catanzaro.
\newblock Few-shot video-to-video synthesis.
\newblock In {\em NeurIPS}, 2019.

\bibitem{vs_2}
Ting-Chun Wang, Ming-Yu Liu, Jun-Yan Zhu, Guilin Liu, Andrew Tao, Jan Kautz, and Bryan Catanzaro.
\newblock Video-to-video synthesis.
\newblock In {\em NeurIPS}, 2018.

\bibitem{uvo}
Weiyao Wang, Matt Feiszli, Heng Wang, and Du Tran.
\newblock Unidentified video objects: A benchmark for dense, open-world segmentation.
\newblock In {\em ICCV}, 2021.

\bibitem{on_4}
Huaxin Xiao, Jiashi Feng, Guosheng Lin, Yu Liu, and Maojun Zhang.
\newblock Monet: Deep motion exploitation for video object segmentation.
\newblock In {\em CVPR}, 2018.

\bibitem{matching_5}
Haozhe Xie, Hongxun Yao, Shangchen Zhou, Shengping Zhang, and Wenxiu Sun.
\newblock Efficient regional memory network for video object segmentation.
\newblock In {\em CVPR}, 2021.

\bibitem{youtube_vos}
Ning Xu, Linjie Yang, Yuchen Fan, Jianchao Yang, Dingcheng Yue, Yuchen Liang, Brian Price, Scott Cohen, and Thomas Huang.
\newblock Youtube-vos: Sequence-to-sequence video object segmentation.
\newblock In {\em ECCV}, 2018.

\bibitem{sam_meets_videos}
Jinyu Yang, Mingqi Gao, Zhe Li, Shang Gao, Fangjing Wang, and Feng Zheng.
\newblock Track anything: Segment anything meets videos.
\newblock {\em arXiv preprint arXiv:2304.11968}, 2023.

\bibitem{on_3}
Linjie Yang, Yanran Wang, Xuehan Xiong, Jianchao Yang, and Aggelos~K Katsaggelos.
\newblock Efficient video object segmentation via network modulation.
\newblock In {\em CVPR}, 2018.

\bibitem{aot}
Zongxin Yang, Yunchao Wei, and Yi Yang.
\newblock Associating objects with transformers for video object segmentation.
\newblock {\em NeurIPS}, 2021.

\bibitem{deaot}
Zongxin Yang and Yi Yang.
\newblock Decoupling features in hierarchical propagation for video object segmentation.
\newblock {\em NeurIPS}, 2022.

\bibitem{v_decomposition_1}
Vickie Ye, Zhengqi Li, Richard Tucker, Angjoo Kanazawa, and Noah Snavely.
\newblock Deformable sprites for unsupervised video decomposition, 2022.

\bibitem{IVOS-W}
Zhaoyuan Yin, Jia Zheng, Weixin Luo, Shenhan Qian, Hanling Zhang, and Shenghua Gao.
\newblock Learning to recommend frame for interactive video object segmentation in the wild.
\newblock In {\em CVPR}, 2021.

\bibitem{zhou17cvpr}
B. Zhou, H. Zhao, X. Puig, S. Fidler, A. Barriuso, and A. Torralba.
\newblock Scene parsing through {ADE20K} dataset.
\newblock In {\em CVPR}, 2017.

\end{thebibliography}
}

\end{document}